# Novel Intrusion Detection using Probabilistic Neural Network and Adaptive Boosting

Tich Phuoc Tran, Longbing Cao
Faculty of Engineering and Information Technology
University of Technology, Sydney, Australia
{tiptran, lbcao}@it.uts.edu.au

Dat Tran
Faculty of Information Sciences and Engineering
University of Canberra, Australia
Dat.Tran@canberra.edu.au

Cuong Duc Nguyen
School of Computer Science and Engineering
International University, HCMC, Vietnam
ndcuong@hcmiu.edu.vn

*Abstract*— **This article applies Machine Learning techniques to solve Intrusion Detection problems within computer networks. Due to complex and dynamic nature of computer networks and hacking techniques, detecting malicious activities remains a challenging task for security experts, that is, currently available defense systems suffer from low detection capability and high number of false alarms. To overcome such performance limitations, we propose a novel Machine Learning algorithm, namely Boosted Subspace Probabilistic Neural Network (BSPNN), which integrates an adaptive boosting technique and a semi-parametric neural network to obtain good trade-off between accuracy and generalty. As the result, learning bias and generalization variance can be significantly minimized. Substantial experiments on KDD-99 intrusion benchmark indicate that our model outperforms other state-of-the-art learning algorithms, with significantly improved detection accuracy, minimal false alarms and relatively small computational complexity.**

## I. Introduction

As more and more corporations rely on computers and networks for communications and critical business transactions, securing digital information has become one of the largest concerns of the business community. A powerful security system is not only a requirement but essential to the livelihood of enterprises. In recent years, there has been a great deal of research conducted in this area to develop intelligent and automated security tools which can fight the latest cyber attacks. Alongside with static defense mechanisms such as keeping operating systems up-to-date or deploying firewalls at critical network segments for access control, more advanced defense systems, namely Intrusion Detection Systems (IDS), are becoming an important part of today's network security architectures. Particularly, IDS can be used to monitor computers or networks for unauthorized activities based on network traffic or system usage behaviors, thereby detect if a system is targeted by a network attack such as a denial of service attack.

The majority of currently existing IDS face a number of challenges such as *low detection rates* which can miss serious intrusion attacks and *high false alarm rates,* which falsely classifies a normal connection as an attack and therefore obstructs legitimate user access to the network resources [1]. These problems are due to the sophistication of the attacks and their intended similarities to normal behavior. More intelligence is brought into IDS by means of Machine Learning (ML). Theoretically, it is possible for a ML algorithm to achieve the best performance, i.e. it can minimize the false alarm rate and maximize the detection accuracy. However, this normally requires infinite training sample sizes (theoretically) [2]. In practice, this condition is impossible due to limited computational power and real-time response requirement of IDS. IDS must be active in real time and they cannot allow much delay because this would cause a bottleneck to the whole network.

To overcome the above limitations of currently existing IDS, we propose an efficient *Boosted Subspace Probabilistic Neural Network* (*BSPNN*) to enhance the performance of intrusion detection for rare and complicated attacks. *BSPNN* combines and improves a Vector Quantized-Generalized Regression Neural Network (VQ-GRNN) with an ensemble technique to improve detection accuracy while minimizing computation overheads by tuning of models. Because this method combines the virtues of boosting and neural network technologies, it has both high data fitting capability and high system robustness. To evaluate our approach, substantial experiments are conducted on the KDD-99 intrusion detection benchmark. The proposed algorithm clearly demonstrates superior classification performance compared with other well-known techniques in terms of bias and variance for the real life problems.

## II. Network Intrusion Detection and Related Works

Because most computers today are connected to the Internet, network security has become a major concern for organizations throughout the world. Alongside the existing techniques for preventing intrusions such as encryption and firewalls, Intrusion Detection technology has established itself as an emerging research field that is concerned with detecting unauthorized access and abuse of computer systems from both internal users and external offenders. An Intrusion Detection System (IDS) is defined as a protection system that monitors computers or networks for unauthorized activities based on network traffic or system usage behaviors, thereby detecting if a system is targeted by a network attack such as a denial of service attack [4]. In response to those identified adversarial transactions, IDS can inform relevant authorities to take corrective actions.

There are a large number of IDS available on the market to complement firewalls and other defense techniques. These systems are categorized into two types of IDS, namely (1) *misuse-based detection* in which events are compared against pre-defined patterns of known attacks and (2) *anomaly-based detection* which relies on detecting the activities deviating from system "normal" operations.

In addition to the overwhelming volume of generated network data, rapidly changing technologies present a great challenge for today's security systems with respect to attack detection speed, accuracy and system adaptability. In order to overcome such limitations, there has been considerable research conducted to apply ML algorithms to achieve a generalization capability from limited training data. That means, given known intrusion signatures, a security system should be able to detect similar or new attacks. Various techniques such as association rules, clustering, Naïve Bayes, Support Vector Machines, Genetic Algorithms, Neural Networks, and others have been developed to detect intrusions. This section provides a brief literature review on these technologies and related frameworks.

One of the rule-based methods which is commonly used by early IDS is the *Expert System* (ES) [3, 4]. In such a system, the knowledge of human experts is encoded into a set of rules. This allows more effective knowledge management than that of a human expert in terms of reproducibility, consistency and completeness in identifying activities that match the defined characteristics of misuse and attacks. However, ES suffers from low flexibility and robustness. Unlike ES, data mining approaches derive association rules and frequent episodes from available sample data, not from human experts. Using these rules, Lee et. al. developed a data mining framework for the purpose of intrusion detection [5, 6]. In particular, system usage behaviors are recorded and analyzed to generate rules which can recognize misuse attacks. The drawback of such frameworks is that they tend to produce a large number of rules and thereby, increase the complexity of the system.

Decision trees are one of the most commonly used supervised learning algorithms in IDS [7-11] due to its simplicity, high detection accuracy and fast adaptation. Another high performing method is Artificial Neural Networks (ANN) which can model both linear and non-linear patterns. ANN-based IDS [12-15] have achieved great successes in detecting difficult attacks. For unsupervised intrusion detection, data clustering methods can be applied [16, 17]. These methods involve computing a distance between numeric features and therefore they cannot easily deal with symbolic attributes, resulting in inaccuracy.

Another well-known ML techniques used in IDS is Naïve Bayes classifiers [7]. Because Naïve Bayes assumes that features are independent, which is often not the case for intrusion detection, correlated features may degrade its performance. In [18], the authors apply a Bayesian network for IDS. The network appears to be attack specific and its size grows rapidly as the number of features and attack types increase.

Beside popular decision trees and ANN, Support Vector Machines (SVMs) are also a good candidate for intrusion detection systems [14, 19] which can provide real-time detection capability, deal with large dimensionality of data. SVMs plot the training vectors in high dimensional feature space through nonlinear mapping and labeling each vector by its class. The data is then classified by determining a set of support vectors, which are members of the set of training inputs that outline a hyperplane in the feature space.

Several other AI paradigms including linear genetic programming [20] , Hidden Markov Model [21], Columbia Model [22] and Layered Conditional Random Fields [23] have been applied for the design of IDS.

III. BOOSTED SUBSPACE PROBABILISTIC NEURAL NETWORK (BSPNN)

*A. Bias-Variance-Computation Dilemma*

Several ML techniques have been adopted in the Network Security domain with certain success; however, there remain severe limitations. Firstly, we consider Artificial Neural Network (ANN) because of its wide popularity and well-known characteristics. As a flexible "model-free" learning method, ANN can fit training data very well and thus provide a low learning bias. However, they are susceptible to overfitting, which can cause instability in generalization [24]. Recent remedies try to improve the model stability by reducing generalization variance at the cost of worse learning bias, i.e. allowing underfitting. However, underfitting is not acceptable for some applications requiring high classification accuracy. Therefore, a system which can achieve both stable generalization and accurate learning is imperative for applications as in Intrusion Detection [19]. Mathematically, both bias and variance may be reduced at the same time given infinite sized models. However, this is infeasible since computing resources must be limited in real life. We develop a learning algorithm which provides a good tradeoff for learning bias, generalization variance and computational requirement motivated by the need of an accurate detection system for Intrusion Detection.

*B. Objectives*

This paper is inspired by a light-weight ANN model, namely Vector Quantized-Generalized Regression Neural Network (VQ-GRNN) [25], which reduces the nonparametric GRNN [26] to a semi-parametric model by applying vector quantization techniques on the training data, i.e. clustering the input space into a smaller subspace. Compared with GRNN method which incorporates every training vector into its structure, VQ-GRNN only applies on a smaller number of clusters of input data. This significantly improves the robustness of the algorithm (low variance), but also controls its learning accuracy to some extent [24]. To make the VQ-GRNN suitable for Intrusion Detection problems, i.e. enhancing its accuracy, we propose the *Boosted Subspace Probabilistic Neural Network* (*BSPNN*) which combines VQ-GRNN and Ensemble Learning technique. Ensemble methods such as *Boosting* [27] iteratively learn multiple classifiers (base classifiers) on different distributions of training data. It particularly guides changes of the training data to direct further classifiers toward more "difficult cases", i.e. putting more weights for previously misclassified instances. It then combines base classifiers in such a way that the composite – boosted learner – outperforms the single classifiers. Amongst popular boosting variants, we choose *Adaptive Boosting* or *AdaBoost* [28] to improve performance of VQ-GRNN. AdaBoost is the most widely adopted method which allows the designer to continue adding weak learners whose accuracy is only moderate until some desired low training error has been achieved. AdaBoost is "adaptive" in the sense that it does not require prior knowledge of the accuracy of these hypotheses [27]. Instead, it measures the accuracy of a base hypothesis at each iteration and sets its parameters accordingly.

Although classifier combinations (as in boosting) can improve generalization performance, correlation between individual classifiers can be harmful to the final composite model. Moreover, it is widely accepted that generalization performance of a combined classifier is not necessarily achieved by combining classifiers with better individual performance but by including independent classifiers in the ensemble [9]. Therefore, such independence condition among individual classifiers which is normally termed as *orthogonality*, *diversity* or *disagreement* is required to obtain a good ensemble.

*C. Model description*

As shown in Figure 1, the proposed *BSPNN* algorithm has two major modules: the *Adaptive Booster* and the *Modified Probabilistic Classifier*.

Given the input data $S = \{(x_i, y_i) | i = 1 \ldots N\}$ where output vector $y \in \{1 \ldots K\}$, the *BSPNN* algorithm aims to produce a classifier $F$ such that:

$$F(x_i) = y_i$$

In this research, we implement $F$ (referred to as *Adaptive Booster*), using SAMME algorithm [29].

$F$ learns by iteratively training a *Modified Probabilistic Classifier f* on weighted data samples $S$ and their weights are updated by the *Distribution Generator* according to previously created models of *f*. This base learner *f* is actually a modified version of the emerging VQ-GRNN model [25] (called *Modified GRNN Base learner*) in which the input data space is reduced significantly (by the *Weighted vector quantization* module) and its output is computed by a linearly weighted mixture of Radial Basis Function (RBF). This process is repeated until $F$ reaches a desired number of iterations or its Mean Squared Error (MSE) approaches an appropriate level. The base hypotheses returned from *f* are finally combined by the *Hypothesis Aggregator*:

$$F(x) = \sum_{m=1}^{M} \alpha_m . f_m(x)$$

This combination depends not only on the misclassification error of previously added $f_m$ but also the diversity $\alpha_m$ of the ensemble at that time. The *Diversity Checker* measures ensemble diversity by using Kohavi-Wolpert variance [30] (which is denoted by the hypothesis weighting coefficient $\alpha_m$). To avoid any confusion, the adaptive booster $F$ is called the *master algorithm* while *f* refers to the *base learner*. They are described in greater details in next sections.

*1) Adaptive Booster*

The Adaptive Booster iteratively produces base hypotheses on a weighted training dataset. The weights are updated adaptively based on the classification performance of component hypotheses. The generated hypotheses are then integrated via a weighted sum based on their diversity.

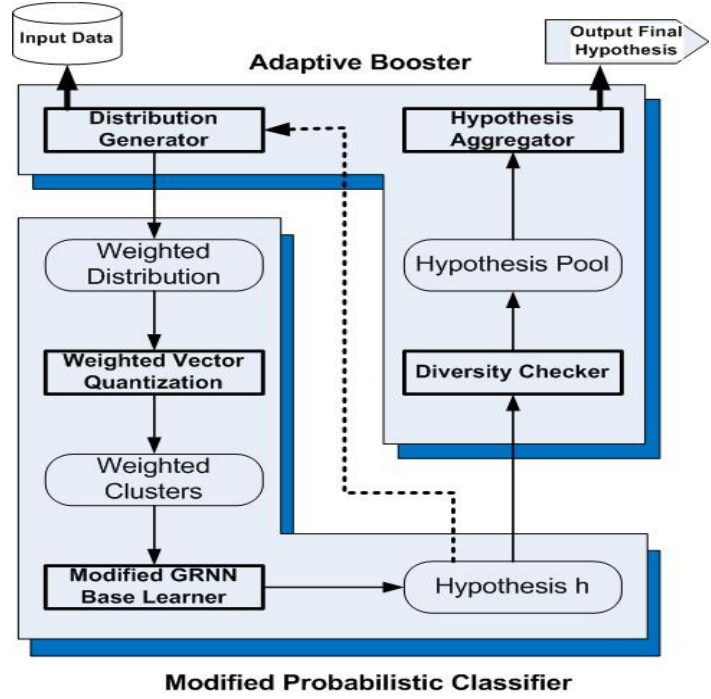

Figure 1. BSPNN high-level design view

TABLE I. ADAPTIVE BOOSTER ALGORITHM

**Input**: $S = \{(x_1, y_1), \ldots, (x_N, y_N)\}$ and associated distribution $W$
**Initialize** $W_i^{(1)} = \frac{1}{N}$ for all $i=1\ldots N$, $\alpha^{(1)} = 1$
**Do for** $t = 1 \ldots T$
**Generate base classifiers**
Train a classifier on the weighed sample $\{S, W^{(t)}\}$ using the Modified Probabilistic Classifier and obtain hypothesis $h^{(t)}: x \to [0,1]^K$
**Compute Kohavi-Wolpert variance ($\alpha^{(t)}$) of current ensemble**
$\alpha^{(t+1)} = \frac{1}{N \cdot L^2} \sum_{j=1}^{N} l(x_j)\left(L - l(x_j)\right)$
Where $L$ and $l(x_j)$ are the number of base classifiers generated so far in the ensemble and the number of classifiers that correctly classifies $x_j$. We have $L=t$.
**Compute class probability estimates**
$C_k^{(t)}(x) = (K - 1) \cdot \left[\log p_k^{(t)}(x) - \frac{1}{K} \sum_{k'=1}^{K} \log p_{k'}^{(t)}(x)\right], k = 1, \ldots, K$
Where $p_k^{(t)}(x) = Prob_w\left(h^{(t)}(x) = k | x\right)$ is the weighted class probability of class $k$.
**Update weights**
$W_i^{(t+1)} = W_i^{(t)} \cdot \exp\left[-\frac{K-1}{K} \cdot \log p^{(t)}(x_i) \cdot h^{(t)}(x_i)\right], i = 1, \ldots, n$
Where $p(x_i) = Prob(x_i)$
**Renormalize** $W$
$W_i = \frac{W_i}{\sum_{j=1}^{N} W_j}, i = 1 \ldots N$
**End for**
**Output**
$C_{final}(x) = \arg\max_k \sum_{t=1}^{T} \alpha^{(t)} \cdot C_k^{(t)}(x)$

*2) Modified Probabilistic Classifier (Base Learner)*

The Modified Probabilistic Classifier serves as the base learner which can be trained on $\{S, d_t\}$ repeatedly by the Adatptive Booster to obtain the hypothesis

$$h_t: x \to [-1, +1]$$

In each boosting iteration, a base hypothesis is created with associated accuracy and diversity measures. From this information, the data weights are updated for the next iteration and the final weighting of that hypothesis in the joint classification is computed.

We adapt VQ-GRNN [25] as a base learner in our *BSPNN* model. VQ-GRNN is closely related to Specht's GRNN [26] and PNN [31] classifiers. This adaptation of VQ-GRNN can produce confidence-rated outputs and it is modified such that it utilizes weights associated with training examples (to compute cluster center vectors and find a single smoothing factor) and incorporates these weights as penalties for misclassifications (e.g. weighted MSE). This modified version of VQ-GRNN is similar to the original one in that a single kernel bandwidth is tuned to achieve satisfactory learning. They both cluster close training vectors according to a very simple procedure related to vector quantization. A number of equally sized radial basis functions are placed at each and every center vector location. These functions are approximated:

$$\sum_{i=0}^{Z_k} f_i(\underline{x} - \underline{x}_i, \delta) \approx Z_k f_k(\underline{x} - \underline{c}_k, \delta)$$

This approximation is reasonable because the $x_i$ vectors are close to each other in the input vector space. Using this idea, the VQ-GRNN's equation can be generalized [25]:

$$\hat{y}(\underline{x}) = \frac{\sum_{i=0}^{M} Z_i y_i f_i(\underline{x} - \underline{c}_i, \delta)}{\sum_{i=0}^{M} Z_i f_i(\underline{x} - \underline{c}_i, \delta)}$$

Where $\underline{c}_i$ is the center vector for class *i* in the input space, $f_i(x, \delta)$ is the radial basis function with centre *x* and the width parameter $\delta$, $y_i$ is the ouput related related to $\underline{c}_i$, $Z_i$ is the number of vectors $x_j$ associated with centre $\underline{c}_i$. $\sum_i Z_i = NV$ is the total number of training vectors.

The above formula can be extended to a multiclass classification problem by redefining the output vector as a *K*-dimensional vector (*K* is the number of classes):

$$y_i = \{y_{i1}, \dots, y_{iK}\}^T$$

where $y_{ik}$ is the class membership probability of the *k*-th class of the vector $x_i$. If the vector $x_i$ is of class *k*, then $y_{ik} = 1.0$ and $y_{ik'} = 0$ for the remaining vector elements ($k \neq k'$). An input vector *x* is classified to class-*k* if the *k*-th element of the output vector has the highest magnitude.

To suit ensemble learning, VQ-GRNN is adapted such that it incoperates the weights associated with each training vector into the learning process, i.e. using them in cluster center formation and Mean Square Error (MSE) calculation for realzing the smoothing factor $\delta$.

Such modifications make VQ-GRNN specially suited for boosting. In particular, the center vector $\underline{c}_i$ is computed as:

$$\underline{c}_k = \frac{\sum_{i=1}^{Z_k} W_i x_i}{Z_k}$$

where $Z_k$ is the number of training vectors belonging to a cluster *k*; $W_i$ is the weight associated with $x_i$.

VQ-GRNN's learning involves finding the optimal bandwidth $\delta$ giving the minimum MSE. In our implementation, a Weighted MSE (WMSE) is used instead:

$$WMSE = \frac{\sum_{i=1}^{N}[W_i(\hat{y}_i(x_i) - y_i)]^2}{N}$$

where $w_i$ and $\hat{y}_i$ are the associated weight and prediction of an example $(x_i, y_i)$, $i = 1\dots N$

*3) Remarks on BSPNN*

The high accuracy of *BSPNN* can be attributed to the boosting effects of SAMME method implemented in the Adaptive Booster module. By sufficiently handling the multiclass problem and using confidence-rated predictions, SAMME can maximize the distribution margins of the training data [32]. Also, our implementation of *Kohavi-Wolpert* variance (*KW*) [30] in the reweighting of hypotheses in the joint classification can effectively enforce the ensemble diversity. The Modified Probabilistic Classifier has very fast adaptation and it is modified to better integrate with the Adaptive Booster module. Particularly, after being modified, it can produce confidence rated outputs and fully utilize the weights given by the booster into learning process. In the next sections, we apply *BSPNN* into specific Intrusion Detection problems.

IV. APPLICATION TO NETWORK INTRUSION DETECTION

Current IDS suffer from low detection accuracy and insufficient system robustness for new and rare security breaches. In this section, we apply our *BSPNN* to identify known and novel attacks in the KDD-99 dataset [1], containing TCP/IP connection records. Each record consisted of 41 attributes (features) and one target value (labeled data) which indicates whether a connection is *Normal* or an attack. There are 40 types of attacks, classified into four major categories, namely Probing (*Probe*) (collect information of target system prior to an attack), Denial of Service (*DoS*) (prevent legitimate requests to a network resource by consuming the bandwidth or overloading computational resources), User-to-Root (*U2R*) (attackers with normal user level access gain privileges of root user), and Remote-to-Local (*R2L*) (unauthorized users gain the ability to execute commands locally).

Table 2 describes the components of KDD-99 dataset (referred to as *Whole KDD*): *10% KDD* containing 26 known attack types (for training) and *Corrected KDD* containing 14 novel attacks (for testing).

TABLE II. KDD-99 COMPONENT DATASETS [1]

| Dataset | DoS | Probe | U2R | R2L | Total Attack | Total Normal |
|---|---|---|---|---|---|---|
| Whole KDD | 3883370 | 41102 | 52 | 1126 | 3925650 | 972780 |
| 10% KDD | 391458 | 4107 | 52 | 1126 | 396743 | 97277 |
| Corrected KDD | 229853 | 4166 | 70 | 16347 | 250436 | 60593 |

### A. Experiment Setup

#### 1) Cost-Sensitive Evaluation

Because an error on a particular class may not be equally serious as errors on other classes, we should consider misclassification cost for intrusion detection. Given a test set, the average cost of a classifier is calculated as below [1]:

$$Cost = \frac{1}{N} \sum_{i=1}^{5} \sum_{j=1}^{5} ConfM(i,j) * CostM(i,j) \quad (4)$$

Where

*N*: total number of connections in the dataset

*ConfM(i,j)*: the entry at row *i*, column *j* in the confusion matrix.

*CostM(i,j)*: the entry at row *i*, column *j* in the cost matrix.

#### 2) Datasets Creation

First, we consider *anomaly detection* where only *normal* connection records are available for training. Any connections that differ from these normal records are classified as "abnormal" without further specifying which attack categories it actually belongs to. For this purpose, we filter all known intrusions from the *10% KDD* to form a pure normal dataset (*Norm*).

For *misuse detection*, we inject the 26 known attacks into *Norm* to classify 14 novel ones. For example, from the *Probe* attacks that appeared in the training set (ipsweep., nmap., portspeep., satan.), we aim to detect unseen *Probe* attacks that were only included in testing data (mscan., saint.). In [33], artificial anomalies are added to the training data to help the learner discover a boundary around the available training data. The method particularly changes the value of one feature of a connection while leaving other features unaltered. However, we do not adopt this method due to its high false alarm rate and its unconfirmed assumption that the boundary is very close to the known data and that they do not intersect one another. Instead, we group 26 known intrusions into 13 clusters $C_1, \dots, C_{13}$ (note that these clusters are not artificially generated but real incidents, available in "10% KDD" set) and use it for classification. Each cluster contains intrusions that require similar features for effective detection and this method, as detailed in [33], is not influenced by cluster orders.

In our experiments, we first created 13 datasets $D_1, \dots, D_{13}$, as shown in Table 3, by incrementally adding each cluster $C_i$ into the normal dataset (*Norm*) to simulate the evolution of new intrusions:

$$D_k = Norm + \sum_{1}^{k} C_i = D_{k-1} + C_k$$

The *BSPNN* and other learning methods are then tested against the "Corrected KDD" testing set, containing both known and unknown attacks.

### B. Experiment Result

#### 1) Anomaly Detection

We train *BSPNN* on the pure Normal dataset (*Norm*) to detect anomalies in "Corrected KDD" testing set. Table 4 shows that our *BSPNN* obtains competitive detection rate compared with [33] while achieves significantly lower false alarm rate (1.1%), minimizing major drawbacks of anomaly detection.

#### 2) Misuse Detection

To test the effect of having known intrusions in the training set on the overall performance, we run *BSPNN* on the 13 training sets: $D_1, \dots, D_{13}$. Its detection rates (*DR*) on different attack categories are displayed in Figure 2. We could discover a general trend of increasing performance as more intrusions are added into training set. In particular, detection of *R2L* attacks requires less known intrusion data (*DR* starts rising at $D_6$) than that of other classes.

Using the full training set ($D_{13}$), we test our *BSPNN* against other existing methods, including the KDD-99 winner [8], the rule-based PNrule approach [34], the multi-class Support Vector Machine [19], the Layered Conditional Random Fields Framework (LCRF) [23], the Columbia Model [22] and the Decision Tree method [11]. Their Detection Rate (*DR*) and False Alarm Rate (*FAR*) are reported in Table 5, with highest *DR* and lowest *FAR* for each class in bold.

TABLE III. CLUSTERS OF KNOWN INTRUSION

| $C_1$ back | $C_2$ buffer_overflow, loadmodule, perl, rootkit |
|---|---|
| $C_3$ ftp_write, warezclient, warezmaster | $C_4$ guess_passwd |
| $C_5$ imap | $C_6$ land |
| $C_7$ portsweep, satan | $C_8$ ipsweep, nmap |
| $C_9$ multihop | $C_{10}$ neptune |
| $C_{11}$ phf | $C_{12}$ pod, teardrop |
| $C_{13}$ spy, smurf | |

TABLE IV. ANOMALY DETECTION RATE (DR) AND FALSE ALARM RATE (FAR) FOR ANOMALY DETECTION

| | Fan et. [33] | *BSPNN* |
|---|---|---|
| Anomaly *DR* | 94.26 | 94.31 |
| *FAR* | 2.02 | 1.12 |

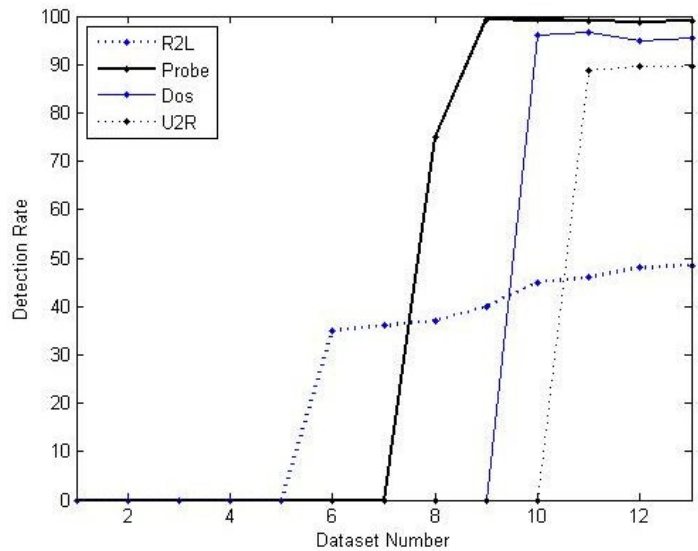

Figure 2. Detection Rate on Datasets for misuse detection

For *Probe* and *DoS* attacks, *BSPNN* can achieve slightly better *DR* than other algorithms with very competitive *FAR*. Though improvement for detection of Normal class is not significant, our model can, in fact, get a remarkably low *FAR*. In addition, a clear performance superiority is claimed for *BSPNN* in the case of *U2R* and *R2L* classes.

It is also important to note that, since KDD-99 dataset is unbalanced (*U2R* and *R2L* appeared rarely), the baseline models can only classify the major classes and performs poorly on other minor ones, while our *BSPNN* exhibits superior detection power for all classes. Significant improvement in detection of more dangerous attacks (*U2R*, *R2L*) leads to lower total weight of misclassification of **0.1523** compared with **0.2332** of the KDD-99 winner.

*(IJCSIS) International Journal of Computer Science and Information Security,*
*Vol. 6, No. 1, 2009*TABLE V. DETECTION RATE (DR) AND FALSE ALARM RATE (FR) FOR MISUSE DETECTION

| | Normal | Probe | DoS | U2R | R2L | DR/FAR (%) |
|---|---|---|---|---|---|---|
| KDD 99 winner [8] | 99.5 | 83.3 | 97.1 | 13.2 | 8.4 | *DR* |
| | 27.0 | 35.2 | 0.1 | 28.6 | 1.2 | *FAR* |
| PNrule [34] | 99.5 | 73.2 | 96.9 | 6.6 | 10.7 | *DR* |
| | 27.0 | 7.5 | **0.05** | 89.5 | 12.0 | *FAR* |
| Multi-class SVM [19] | 99.6 | 75 | 96.8 | 5.3 | 4.2 | *DR* |
| | 27.8 | 11.7 | 0.1 | 47.8 | 35.4 | *FAR* |
| Layered Conditional Random Fields [23] | - | 98.60 | 97.40 | 86.30 | 29.60 | *DR* |
| | - | **0.91** | 0.07 | 0.05 | 0.35 | *FAR* |
| Columbia Model [22] | - | 96.7 | 24.3 | 81.8 | 5.9 | *DR* |
| Decision Tree [11] | - | 81.4 | 60.0 | 58.8 | 24.2 | *DR* |
| ***BSPNN*** | **99.8** | **99.3** | **98.1** | **89.7** | **48.2** | *DR* |
| | **3.6** | **1.1** | **0.06** | **0.03** | **0.19** | *FAR* |

## V. CONCLUSION

This research is inspired by the need of a highly performing but low in computation classifier for applications in Network Security. Particularly, the *Boosted Subspace Probabilistic Neural Network* (*BSPNN*) is proposed which combines two emerging algorithms, an adaptive boosting method and a probabilistic neural network. *BSPNN* retains the semi-parametric characteristics of VQ-GRNN and therefore obtains low generalization variance while receives accuracy boosting from SAMME method (low bias). Though *BSPNN* requires more processing power due to the effect of boosting, the increased computation is still lower than GRNN or other boosted algorithms.

Experiments on the KDD-99 network intrusion dataset show that our approach obtains superior performance in comparison with other state-of-the-art detection methods, achieving low learning bias and improved generalization at an affordable computational cost.

## REFERENCES

[1] C. Elkan, "Results of the KDD'99 Classifier Learning," *ACM SIGKDD Explorations,* vol. 1, pp. 63-64, 2000.

[2] I. Kononenko and M. Kukar, Machine Learning and Data Mining: Introduction to Principles and Algorithms Horwood Publishing Limited, 2007.

[3] D. S. Bauer and M. E. Koblentz, "NIDX – an expert system for real-time network intrusion detection," in *Proceeding of the Computer Networking Symposium* Washington, D.C., 1988, pp. 98-106.

[4] K. Ilgun, R. Kemmerer, and P. Porras, "State transition analysis: a rule-based intrusion detection approach," *IEEE Transactions on Software Engineering,* pp. 181-199, 1995.

[5] W. Lee, S. Stolfo, and K. Mok, "Mining Audit Data to Build Intrusion Detection Models," *Proc. Fourth International Conference Knowledge Discovery and Data Mining* pp. 66-72, 1999.

[6] W. Lee, S. Stolfo, and K. Mok, "A Data Mining Framework for Building Intrusion Detection Model," *Proc. IEEE Symp. Security and Privacy,* pp. 120-132, 1999.

[7] N. B. Amor, S. Benferhat, and Z. Elouedi, "Naive Bayes vs. Decision Trees in Intrusion Detection Systems," *Proc. ACM Symp. Applied Computing,* pp. 420-424, 2004.

[8] B. Pfahringer, "Winning the KDD99 Classification Cup: Bagged Boosting," *SIGKDD Explorations,* vol. 1, pp. 65–66, 2000.

[9] V. Miheev, A. Vopilov, and I. Shabalin, "The MP13 Approach to the KDD'99 Classifier Learning Contest," *SIGKDD Explorations,* vol. 1, pp. 76–77, 2000.

[10] I. Levin, "KDD-99 Classifier Learning Contest: LLSoft's Results Overview," *SIGKDD Explorations,* vol. 1, pp. 67–75, 2000.

[11] J.-H. Lee, J.-H. Lee, S.-G. Sohn, J.-H. Ryu, and T.-M. Chung, "Effective Value of Decision Tree with KDD 99 Intrusion Detection Datasets for Intrusion Detection System," in *10th International Conference on Advanced Communication Technology*. vol. 2, 2008, pp. 1170-1175.

[12] Z. Zhang, J. Li, C. N. Manikopoulos, J. Jorgenson, and J. Ucles, "HIDE: A Hierarchical Network Intrusion Detection System Using Statistical Preprocessing and Neural Network Classification," *Proc. IEEE Workshop Information Assurance and Security,* pp. 85-90, 2001.

[13] J. Cannady, "Artificial neural networks for misuse detection," in *In Proceedings of the National Information Systems Security Conference* Arlington, VA, 1998.

[14] S. Mukkamala, G. Janoski, and A. Sung "Intrusion detection using neural networks and support vector machines," in *International Joint Conference on Neural Networks (IJCNN)*. vol. 2: IEEE, 2002, pp. 1702-1707.

[15] C. Jirapummin, N. Wattanapongsakorn, and P. Kanthamanon, "Hybrid neural networks for intrusion detection system," *In Proceedings of The 2002 International Technical Conference On Circuits/Systems,Computers and Communications,* 2002.

[16] L. Portnoy, E. Eskin, and S. Stolfo, "Intrusion Detection with Unlabeled Data Using Clustering," *Proc. ACM Workshop Data Mining Applied to Security (DMSA),* 2001.

[17] H. Shah, J. Undercoffer, and A. Joshi, "Fuzzy Clustering for Intrusion Detection," *Proc. 12th IEEE International Conference Fuzzy Systems (FUZZ-IEEE '03),* vol. 2, pp. 1274-1278, 2003.90
http://sites.google.com/site/ijcsis/
ISSN 1947-5500

*(IJCSIS) International Journal of Computer Science and Information Security,*
*Vol. 6, No. 1, 2009*

[18] C. Kruegel, D. Mutz, W. Robertson, and F. Valeur, "Bayesian Event Classification for Intrusion Detection," *Proc. 19th Annual Computer Security Applications Conference,* pp. 14-23, 2003.

[19] T. Ambwani, "Multi class support vector machine implementation to intrusion detection," in *Proc. of IJCNN*, 2003, pp. 2300-2305.

[20] D. Song, M. I. Heywood, and A. N. Zincir-Heywood, "Training Genetic Programming on Half a Million Patterns: An Example from Anomaly Detection," *IEEE Trans. Evolutionary Computation,* vol. 9, pp. 225-239, 2005.

[21] W. Wang, X. H. Guan, and X. L. Zhang, "Modeling Program Behaviors by Hidden Markov Models for Intrusion Detection," *Proc. International Conference Machine Learning and Cybernetics,* vol. 5, pp. 2830-2835, 2004.

[22] W. Lee and S. Stolfo, "A Framework for Constructing Features and Models for Intrusion Detection Systems," *Information and System Security,* vol. 4, pp. 227-261, 2000.

[23] K. K. Gupta, B. Nath, and R. Kotagiri, "Layered Approach using Conditional Random Fields for Intrusion Detection," *IEEE Transactions on Dependable and Secure Computing,* vol. 5, 2008.

[24] A. Zaknich, *Neural Networks for Intelligent Signal Processing*. Sydney: World Scientific Publishing, 2003.

[25] A. Zaknich, "Introduction to the modified probabilistic neural network for general signal processing applications," *IEEE Transactions on Signal Processing,* vol. 46, pp. 1980-1990, 1998.

[26] D. F. Spetch, "A general regression neural network," *IEEE Transactions on Neural Networks,* vol. 2, pp. 568-576, 1991.

[27] R. E. Schapire, "A brief introduction to boosting," in *Proceedings of the Sixteenth International Joint Conference on Artificial Intelligence*, San Francisco, CA, 1999, pp. 1401-1406.

[28] Y. Freund and R. Schapire, "A decision-theoretic generation of on-line learning and an application to boosting," *Journal of Computer and System Science,* vol. 55, pp. 119–139, 1997.

[29] J. Zhu, S. Rosset, H. Zhou, and T. Hastie, "Multiclass adaboost," *The Annals of Applied Statistics,* vol. 2, pp. 1290--1306., 2005.

[30] R. Kohavi and D. Wolpert, "Bias plus variance decomposition for zero-one loss functions," in *Proc. of International Conference on Machine Learning* Italy, 1996, pp. 275-283.

[31] D. F. Specht, "Probabilistic neural networks," *Neural Networks,* vol. 3, pp. 109-118, 1990.

[32] J. Huang, S. Ertekin, Y. Song, H. Zha, and C. L. Giles, "Efficient Multiclass Boosting Classification with Active Learning," *ICDM,* 2007.

[33] W. Fan, M. Miller, S. Stolfo, W. Lee, and P. Chan, "Using artificial anomalies to detect unknown and known network intrusions," *Knowledge and Information Systems,* vol. 6, pp. 507–527, 2004.

[34] R. Agarwal and M. V. Joshi, "PNrule: A New Framework for Learning Classifier Models in Data Mining," in *A Case-Study in Network Intrusion Detection*, 2000.

91  http://sites.google.com/site/ijcsis/
ISSN 1947-5500